\pdfoutput=1
\documentclass{article}

\usepackage[preprint]{neurips_2024}

\usepackage{fontawesome}
\usepackage[utf8]{inputenc} 
\usepackage[T1]{fontenc}    
\usepackage{hyperref}       
\usepackage{url}            
\usepackage{booktabs}       
\usepackage{amsfonts}       
\usepackage{nicefrac}       
\usepackage{microtype}      
\usepackage{xcolor}         
\usepackage{amsmath}
\usepackage{amssymb}
\usepackage{nomencl}
\usepackage{glossaries}
\usepackage{xspace}
\usepackage{multirow}
\usepackage{graphicx}
\usepackage{bbding}
\usepackage{pifont}
\usepackage{algorithm}
\usepackage{subfigure}
\usepackage{enumitem}
\usepackage{multicol}
\usepackage{caption}
\usepackage{array}
\usepackage{fp}
\usepackage{makecell}
\usepackage{flushend}
\usepackage{cases}
\usepackage{color}
\usepackage{algpseudocode}
\usepackage{listings}
\usepackage{mathrsfs}
\usepackage{longtable}
\usepackage{colortbl}
\usepackage{bm}
\usepackage{tabularx}
\usepackage{caption}
\setlist[itemize]{noitemsep, topsep=0pt}
\definecolor{codegreen}{rgb}{0,0.3,0.6}
\definecolor{codegray}{rgb}{0.5,0.5,0.5}

\newcommand{\ie}{\emph{i.e.,}\xspace}
\newcommand{\eg}{\emph{e.g.,}\xspace}

\newcommand{\paratitle}[1]{\vspace{1.5ex}\noindent\textbf{#1}}

\newcommand{\ignore}[1]{}

\usepackage{tcolorbox}
\definecolor{darkorange}{RGB}{255, 140, 0}
\definecolor{lightgreen}{RGB}{145, 204, 117}
\definecolor{lightyellow}{RGB}{250, 200, 88}
\definecolor{lightred}{RGB}{238, 102, 102}
\definecolor{lightblue}{RGB}{115, 192, 222}
\newtcolorbox{promptbox}[2][Prompt]{
colback=black!5!white,
arc=5pt, 
boxrule=0.5pt,
fonttitle=\bfseries,
title=#1, 
before upper={\scriptsize}, fontupper=\fontfamily{ptm}\selectfont,
colframe=#2, 
}

\title{Towards Effective Code-Integrated Reasoning}

\author{%
 Fei Bai$^{1}$\thanks{Equal contribution.}~,
  Yingqian Min$^{1*}$,
  Beichen Zhang$^{1*}$,
  Zhipeng Chen$^{1}$\\
  \textbf{Wayne Xin Zhao$^{1}$\thanks{Correspondence to Wayne Xin Zhao.}, ~Lei Fang$^3$, ~Zheng Liu$^2$, ~Zhongyuan Wang$^2$, ~Ji-Rong Wen$^1$}
  \vspace{3pt} \\
  $^1$Gaoling School of Artificial Intelligence, Renmin University of China\\
  $^2$BAAI~~~~~
  $^3$DataCanvas Alaya NeW\\
  {\small\texttt{\{yingqianm,zhipeng\_chen,jrwen\}@ruc.edu.cn}}\\
  {\small\texttt{\{feibaienoch,zhangbeichen724,batmanfly\}@gmail.com}}
}

\begin{document}
\textit{Technical Report on Slow Thinking with LLMs: Code-Integrated Reasoning}

\maketitle

\begin{abstract}
In this paper, we investigate \emph{code-integrated reasoning}, where models generate code when necessary and integrate feedback by executing it through a code interpreter. To acquire this capability, models must learn when and how to use external code tools effectively, which is supported by tool-augmented reinforcement learning (RL) through interactive learning. Despite its benefits, tool-augmented RL can still suffer from potential instability in the learning dynamics.
In light of this challenge, we present a systematic approach to improving the training effectiveness and stability of tool-augmented RL for code-integrated reasoning. Specifically, we develop enhanced training strategies that balance exploration and stability, progressively building tool-use capabilities while improving reasoning performance.
Through extensive experiments on five mainstream mathematical reasoning benchmarks, our model demonstrates significant performance improvements over multiple competitive baselines. 
Furthermore, we conduct an in-depth analysis of the mechanism and effect of code-integrated reasoning, revealing several key insights, such as the extension of model’s capability boundaries and the simultaneous improvement of reasoning efficiency through code integration. All data and code for reproducing this work are available at: \url{https://github.com/RUCAIBox/CIR}.

\end{abstract}

\section{Introduction}
\label{sec-intro}
Recent advances in large reasoning models (\eg DeepSeek-R1~\cite{deepseekr1}) have demonstrated significant performance gains through increased test-time computation. Unlike conventional language models, these models do not produce answers directly. Instead, they engage in a deliberative thought process—searching, proposing, verifying, and evaluating potential solutions to arrive at the correct answer. 
However, even with more sophisticated reasoning procedures, their capabilities remain fundamentally constrained by the inherent limitations of large language models~\cite{zhao2023survey,yue2025does}.
Since they still rely on the same underlying training and generation mechanisms, longstanding issues, such as imprecise numerical computation and restricted knowledge coverage, persist, ultimately diminishing models' effectiveness.

To overcome these inherent constraints, augmenting LLMs with external tools has emerged as a promising solution. For instance, generating and executing code during reasoning has proven effective in enhancing mathematical problem-solving, while integrating search engines can compensate for limited or outdated knowledge.
Building on these insights~\cite{chen2025empirical,li2025torl,feng2025retool,mai2025agent,song2025r1,sun2025simpledeepsearcher}, numerous studies have explored the integration of external tools into reasoning models. The predominant approach aims at encouraging LLMs to invoke appropriate tools during reasoning. Typically, models are either fine-tuned using supervised demonstrations or directly optimized through reinforcement learning~(RL) with specialized reward models designed for tool usage.
RL-based methods have demonstrated particular promise by intrinsically incentivize LLMs to use tools effectively, which are often termed as \emph{tool-augmented RL}.

However, tool-augmented RL methods present distinct challenges compared to standard RL approaches for reasoning. 
First, the training process becomes more intricate as models must learn both reasoning capabilities and proper tool utilization. Furthermore, external tool invocation can interrupt the reasoning sequence by introducing non-reasoning tokens, potentially destabilizing the training process. Additionally, these approaches further exacerbate the already intensive computational demands of reinforcement learning, as they require frequent tool invocation and increased interaction rounds during training. Although prior work~\cite{wang2025ragen,song2025r1-2} has identified and partially addressed these issues, comprehensive solutions that systematically consider these factors beyond basic effectiveness remain largely unexplored.

In this paper, we investigate the integration of a specialized tool—a code interpreter—to enhance reasoning model performance. Unlike conventional tools such as search engines or calculators, reasoning models must generate complex code snippets that directly participate in the reasoning process, rather than simply producing search queries or mathematical expressions. This approach fundamentally depends on systematic integration between natural language and formal programming languages to achieve improved reasoning capabilities. We define this reasoning paradigm as \emph{code-integrated reasoning}.

To address the training challenge in tool-augmented RL, we introduce enhanced RL strategies that balance exploration and stability objectives for code-integrated reasoning. For exploration, we remove the KL divergence term and adopt the clip-higher method inspired by DAPO~\cite{yu2025dapo}, while progressively increasing the tool interaction budget. To ensure stability, we set the entropy coefficient to zero, enforce precise code block matching to minimize rollout noise, and mask feedback from external tools. This dual-focused approach achieves state-of-the-art performance across multiple mainstream benchmarks, outperforming competitive baselines.

In addition to technical improvements, we analyze how and when code integration enhances the reasoning performance of language models. First, code integration significantly extends the capacity boundaries of reasoning models (measured as \textsc{Pass@k}) by leveraging the computational capabilities of code interpreters. Second, unlike long-chain-of-thought (long-CoT) reasoning, code-integrated reasoning produces more concise and efficient reasoning paths, typically beginning with a brief yet complete solution overview before generating executable code. Third, while executable but logically flawed code can hinder model performance, non-executable code may still contribute to correct solutions by compelling the model to reflect on and revise its output. Finally, the benefits of code-integrated reasoning vary across problem types, with geometry problems showing minimal improvement among four selected types.

To facilitate the reproduction of our work, we release the datasets, code and model checkpoints at the link: \url{https://github.com/RUCAIBox/CIR}, which will be useful for subsequent research. 

\section{Methodology}
\label{sec-method}

In this section, we present our proposed approach for code-integrated reasoning, beginning with a formulation of this reasoning paradigm, followed by an analysis of training challenges and corresponding improvement strategies.

\subsection{Code-Integrated Reasoning}
We first formalize the \emph{code-integrated reasoning} setting for our task, which empowers large language models (LLMs) to tackle complex reasoning tasks by adaptively incorporating intermediate code execution during inference. In this paradigm, code generation is selectively triggered during inference—typically when the model encounters scenarios that benefit from precise computation, such as numerical operations, logical conditions, or symbolic transformations.

Given a problem \( q \), the model generates code snippets conditioned on both \( q \) and the current reasoning context \( h_t \) when code execution is triggered. These snippets are then executed by an external interpreter, and their results are reintegrated into the reasoning sequence to guide subsequent steps. 
This iterative loop of code generation, execution, and context update allows the model to refine its reasoning steps and generate more accurate and verifiable outputs. The process is formally described as follows:

\begin{align}
c_t &= f_\theta(q, h_t), \label{eq:code_gen} \\
r_t &= \mathcal{I}(c_t), \label{eq:code_exec} \\
h_{t+1} &= h_t \cup c_t \cup r_t, \label{eq:context_update}
\end{align}
where $t$ denotes   the $t$-th code generation step for solving problem $q$.  
The reasoning process starts from \( h_0 \) and advances step by step, ultimately yielding the final answer \( y \): 



\begin{equation}
(q, h_0) \xrightarrow{f_\theta} c_0 \xrightarrow{\mathcal{I}} r_0 \rightarrow h_1 \xrightarrow{f_\theta} \cdots \xrightarrow{f_\theta} c_T \xrightarrow{\mathcal{I}} r_T \rightarrow h_{T+1} \xrightarrow{f_\theta} y,
\label{eq:full_process}
\end{equation}

where \( q \) is the input problem, and \( h_t \) represents the model's generated sequence up to step \( t \), including all intermediate codes and corresponding results. The function \( f_\theta \), parameterized by \( \theta \), generates the code snippet \( c_t \), which is executed by the external interpreter \( \mathcal{I} \) to yield the result \( r_t \). This result is appended to the sequence, producing the updated context \( h_{t+1} \). The final answer \( y \) is generated based on the complete reasoning trajectory.

A key distinction between code-integrated reasoning and other tool-augmented approaches  (\eg reasoning with search engines) is that the generated code snippets serve as a formal representation of reasoning steps. This paradigm effectively bridges natural language reasoning with formal reasoning, making it particularly well-suited for tasks demanding algorithmic precision and symbolic manipulation.


\subsection{Tool-Augmented RL Framework}


To develop our code-integrated reasoning model, we begin with base models (\eg Qwen2.5-Math-7B) and employ typical tool-augmented reinforcement learning~(RL) for optimization. 

Conventional RL algorithms like PPO optimize LLMs by maximizing the following surrogate objective: 
\begin{equation}
\mathcal{L}(\theta) = 
\mathbb{E}_{q \sim P(Q), o \sim \pi_{\theta_{\text{old}}}(O|q)} \left[
\frac{1}{|o|} \sum_{t=1}^{|o|} 
\min \left( 
s_t(\theta) A_t, \ 
\text{clip}(s_t(\theta), 1 - \epsilon, 1 + \epsilon) A_t 
\right)
\right]
\label{equation:obj}
\end{equation}
where $s_t(\theta)$ represents the probability ratio between the new policy and old policy for observation $o_t$:
\begin{equation}
s_t(\theta) = \frac{\pi_{\theta}(o_t | q, o_{<t})}{\pi_{\theta_{\text{old}}}(o_t | q, o_{<t})}. 
\end{equation}

The trajectory $o = [o_0, o_1, ..., o_T]$ reflects only the model’s autoregressively generated intrinsic knowledge, excluding any external feedback.

In contrast, the tool-augmented RL framework allows the model to interact with external tools over multiple rounds during rollouts. This process produces trajectories containing detailed interaction information, which are then used to calculate RL experiences, guiding the model to update. Therefore, $s_t(\theta)$ can be modified as:

\begin{equation}
s_t^{'}(\theta) = \frac{\pi_{\theta}(o_t | q, o_{<t};\mathcal{I})}{\pi_{\theta_{\text{old}}}(o_t | q, o_{<t};\mathcal{I})}
\end{equation}

where $\mathcal{I}$ represents the external tool, such as code interpreter. Through integrating feedback from external tools into the trajectory, the model can adjust its decision-making process based on concrete execution results, improving the effectiveness of learning.

\subsection{RL with Exploration-Stability Balance}
In this part, we first discuss the potential instability issues arising from tool-augmented RL during optimization, then present corresponding mitigation strategies. 

\subsubsection{Exploration-Stability Trade-off Issues in Tool-Augmented RL}
Our empirical studies reveal that integrating external tools into reinforcement learning (RL) can introduce instability issues stemming from three primary factors: interaction boundary disruptions, distributional shifts between model reasoning and external feedback, and response homogenization due to fixed interaction budgets. These factors not only complicate the training process but also compromise both learning stability and effectiveness. We next discuss the three factors in detail.



\begin{figure} 
\centering 
\subfigure[Boundary based on stop token.]{\label{fig:boundry1}
\includegraphics[width=0.45\linewidth]{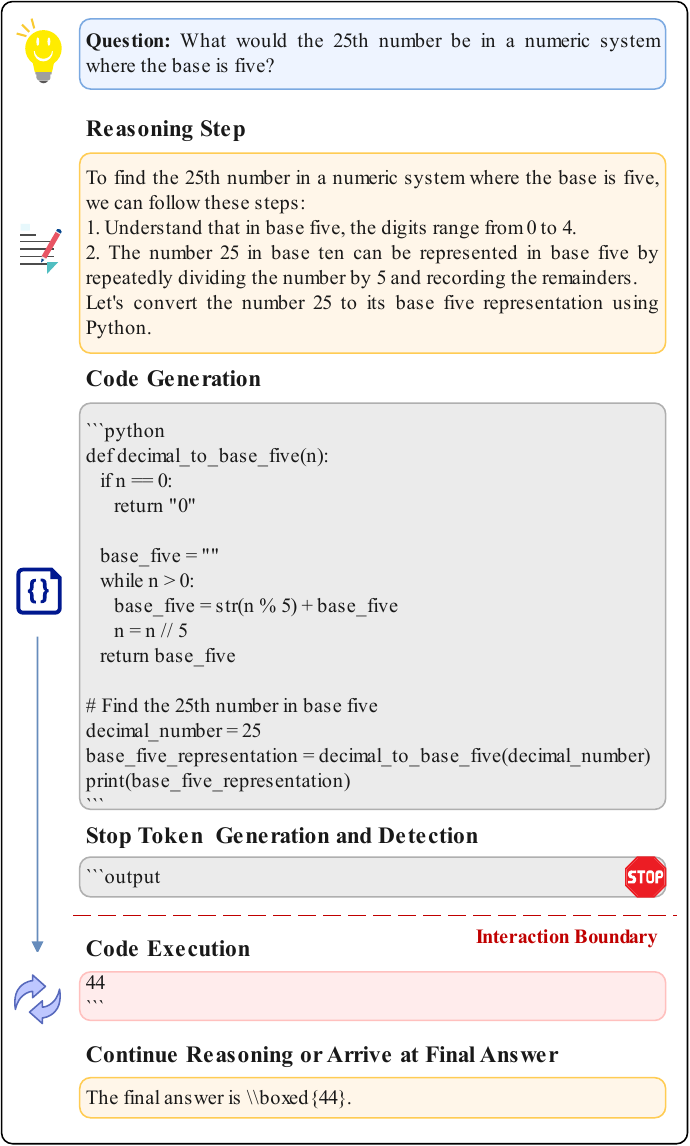}}
\subfigure[Boundary based on precise matching.]{\label{fig:boundry2}
\includegraphics[width=0.45\linewidth]{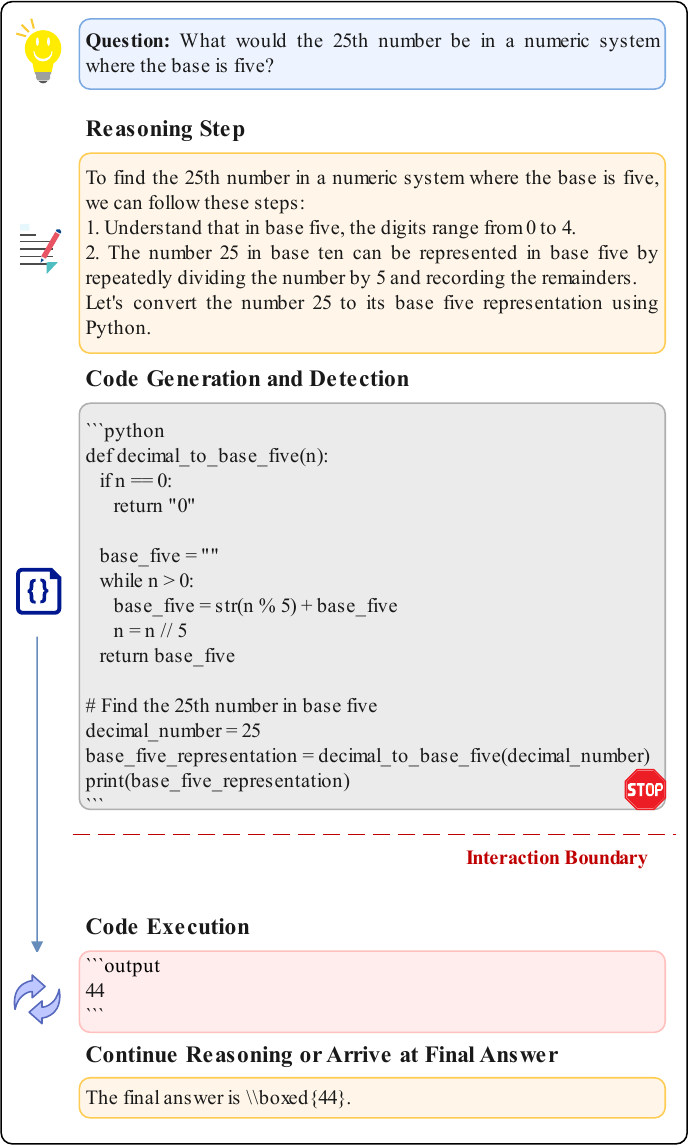}}
\caption{
Illustration of interaction boundary based on stop tokens and precise matching. (a) The model first generates reasoning steps and then generates code. Interaction with the code interpreter is only triggered once a designated stop token, such as \texttt{```}output, is emitted. Upon detecting this token, the preceding code segment is extracted, executed by the code interpreter, and the resulting output is appended to the model’s response, continuing the reasoning process. If the model generates code but fails to immediately emit the stop token, it may introduce noise or even miss a necessary interaction.
(b) Execution is triggered when the model detects a complete and well-formed code block (e.g., \texttt{```}python ... \texttt{```}). At this point, the model pauses its reasoning and interacts with the code interpreter. This exact-match criterion helps ensure that only valid code blocks are executed, preventing noise or omissions from malformed outputs.}
\label{fig:boundry_comparison}
\end{figure}

\paratitle{Instability caused by interaction boundary disruptions.}
To integrate external tools effectively, we must define appropriate \emph{interaction boundaries} that specify both how tool interactions are invoked and how their feedback is incorporated into continued inference. Figure~\ref{fig:boundry1} illustrates this interaction boundary in typical implementations. As shown, this process necessarily involves manually defined boundary conditions, including when model-tool interactions begin and how feedback is integrated into the existing reasoning sequence.
The design of these interaction boundaries directly impacts model training performance. Poorly configured boundaries may even degrade overall model performance.




\paratitle{Impact of distributional shifts on learning.}
Distributional shifts arise between the model's own reasoning process and the feedback returned by external tools, disrupting the continuity of inference. The feedback might differ from the model’s original reasoning patterns, thus interrupting the model’s inference flow. As tool use becomes more frequent, these shifts accumulate over time, gradually undermining learning stability and increasing the risk of training collapse.

\paratitle{Response homogenization under fixed interaction budgets.}
In tool-augmented RL, the risk of response homogenization, commonly known as mode collapse, becomes more pronounced. Due to the fixed tool interaction rounds during training, the model tends to converge on a narrow range of behaviors that are consistently rewarded within the constrained interaction budget, which results in increasingly deterministic outputs, reducing the diversity of reasoning patterns explored by the model.

\subsubsection{Improved Training Strategies}
\label{strategy}

To address the training issues posed by tool-augmented RL, such as instability, distributional shifts, and reduced exploration, we adopt targeted training strategies that balance exploration with stability, enabling more robust learning and improved performance in tool-augmented settings. These strategies can be categorized into two types based on their objectives: (1) \emph{stability maintenance} and (2) \emph{exploration enhancement}.


\paratitle{Stability maintenance.} 
This category of strategies focus on stabilizing the model's learning dynamics, preventing training collapse, and ensuring steady adaptation throughout the learning process.



$\bullet$ \emph{Precise matching of interaction boundaries.}
To mitigate the instability caused by inconsistent tool interaction boundaries, we employ a exact match-based approach (shown in Figure~\ref{fig:boundry2}) to precisely identify and delimit code blocks. In contrast, heuristic methods such as stop tokens are prone to premature termination or inaccurate boundary detection, introducing noise into the training process. By using precise pattern matching, we ensure that the model interacts with well-structured and consistently formatted code segments, reducing ambiguity and improving the reliability of feedback integration.

$\bullet$ \emph{External tool feedback masking.} 
During model optimization, we mask the feedback from external tools when calculating the training loss to prevent the model from learning unstable and variable outputs. This reduces distribution mismatch between the model’s own reasoning and external feedback, allowing the model to focus on critical reasoning patterns and enhancing training stability.

$\bullet$ \emph{Entropy bonus disabling.} 
The entropy bonus encourages exploration in RL, but as training progresses, it often causes policy entropy to increase continuously, resulting in higher behavioral uncertainty. In environments with external tool interactions, increasing randomness amplifies feedback noise, which can destabilize training or cause collapse. Disabling the entropy bonus therefore prevents uncontrolled entropy growth and promotes more stable learning.

\paratitle{Exploration enhancement.}
This category of strategies focuses on encouraging diverse exploration, preventing premature convergence, and improving training effectiveness.


$\bullet$ \emph{Progressive increase of interaction budget.} 
Fixed interaction budgets can lead the model to converge early during training. To address this, we gradually increase the number of tool invocations allowed throughout the training process. This progressive scheduling strategy encourages the model to explore a broader range of reasoning paths in the early stages rather than settling too rapidly on narrowly optimized, high-reward behaviors. By delaying over-stabilization of the policy, the model maintains output diversity for a longer period, fostering more robust strategy learning.

$\bullet$ \emph{KL term removal.} 
The KL divergence term is often applied to keep the updated policy close to a reference policy, preventing drastic changes. In interactive scenarios involving external tools, this constraint can hinder the model’s ability to explore new behaviors, leading to reduced diversity and early convergence. Eliminating the KL term relaxes these restrictions, enabling the model to adopt a wider variety of reasoning strategies and maintain richer exploration throughout training, which enhances overall learning effectiveness.

$\bullet$ \emph{Clip higher.} 
Following DAPO~\cite{yu2025dapo}, we increase the upper bound in the surrogate loss to form an asymmetric clipping strategy. This enhances exploration and preserves policy entropy, encouraging the model to explore a wider range of behaviors in interactive settings.


\section{Experiments}
\label{sec-experimens}
In this section, we first detail the experimental setup and then discuss the results.

\subsection{Experiment Settings}

\paratitle{Training.}
We employ two backbone models for our experiments: Qwen2.5-Math-7B~\cite{Yang2024qwen2.5math} (hereafter \emph{Qwen2.5-M7}), serving as our primary test model, and Qwen3-8B~\cite{yang2025qwen3} in non-thinking mode (denoted as \emph{Qwen3-NT8}).
All training experiments are conducted within the veRL framework~\cite{sheng2024hybridflow}. 
For training data, we use a subset of 38000 samples curated by STILL-3~\cite{chen2025empirical}, which ensures a balanced distribution of problem difficulty. To improve training stability, we employ the REINFORCE++ algorithm~\cite{hu2025reinforce++} alongside the strategies detailed in Section~\ref{strategy}—notably, a progressive increase in tool interaction rounds from 2 to 4. During training, we adopt a binary reward scheme where the model receives a reward of 1 for a correct answer and -1 otherwise. 


\paratitle{Evaluation.} 
To evaluate the effectiveness of our framework, we conduct experiments on five widely used benchmarks: MATH500~\cite{dan2021math}, AMC23, AIME2024, AIME2025 and OlymMATH~\cite{sun2025challenging}. MATH500 comprises 500 competition-level math problems drawn from the MATH test set; AMC23 contains 40 problems aimed at middle and high school students, covering a broad spectrum of mathematical skills and knowledge; AIME2024 and AIME2025 each include 30 problems crafted to assess the advanced problem-solving abilities of top high school students; OlymMATH is an Olympiad-level benchmark intended to rigorously evaluate the complex reasoning capabilities of LLMs, featuring 100 challenging problems in English. To ensure evaluation reliability, we use \textsc{avg@16} (average accuracy over 16 samples, temperature=1.0) for AMC23, AIME2024, AIME2025 and OlymMATH. For MATH500, due to its larger scale and relative stability, we adopt greedy decoding instead.

\paragraph{Baselines.}
We compare our model against a set of competitive baselines, which can be grouped into two categories based on whether code execution is integrated during reasoning. The first group includes models that rely purely on text-based reasoning without code usage. \textsc{Qwen2.5-Math-7B-Instruct} is a strong supervised model obtained through a series of post-training steps on Qwen2.5-Math-7B. \textsc{SimpleRL-Zero-7B}~\cite{zeng2025simplerl}, \textsc{Eurus-2-7B-PRIME}~\cite{cui2025process}, and \textsc{Oat-Zero-7B}~\cite{liu2025understanding} are high-performing models based on either Qwen-Math-7B or Qwen-7B-Base, further optimized from standard reinforcement learning. The second group includes models equipped with code-integrated capabilities during reasoning. \textsc{Qwen2.5-Math-7B-Instruct-TIR} directly integrates code usage at inference time, while \textsc{ToRL-7B}~\cite{li2025torl} is trained via tool-augmented reinforcement learning, allowing dynamic code invocation throughout the training process. This distinction allows us to isolate the effect of code-integrated reasoning and assess its impact across different model designs.

\subsection{Performance Analysis}
After describing the experimental setup, we next present the performance analysis. 

\subsubsection{Main Results}

\begin{table}[htbp]
    \centering
    \small
    \setlength\tabcolsep{2.4pt}
    \caption{Performance of \textsc{CIR} and other baselines. For MATH500 , we utilize greedy decoding strategy for evaluation. For the other four benchmarks, we sample 16 responses and report the \textsc{avg@16} accuracy.}
      \begin{tabular}{lccccccc}
      \toprule
      \textbf{Models} & \textbf{Code} & \textbf{MATH500} & \textbf{AMC23} & \textbf{AIME24} & \textbf{AIME25} &  \textbf{OlymMATH} & \textbf{Avg.} \\
      \midrule
      \textsc{Qwen2.5-Math-7B-Instruct} & \ding{55} & 75.0 & 45.2 & 8.12 & 5.00 & 3.38 & 27.3 \\
      \textsc{Qwen2.5-Math-7B-Instruct-TIR} & \ding{52} & 78.8 & 59.3 & 25.8 & 18.8 & 18.4 & 40.2\\
      \textsc{SimpleRL-Zero-7B} & \ding{55} & 75.4 & 51.1 & 18.3 & 6.46 & 8.75 & 32.0 \\
      \textsc{Eurus-2-7B-PRIME} & \ding{55} & 81.2 & 62.8 & 18.8 & 15.4 & 9.94 & 37.6 \\
      \textsc{Oat-Zero-7B} & \ding{55} & 79.2 & 64.2 & 31.3 & 10.4 & 11.3 & 39.3 \\
      \textsc{ToRL-7B} & \ding{52} & 83.8 & 73.2 & 37.9 & 29.2 & 28.5 & 50.5 \\
      \midrule
      \textsc{CIR~(Qwen2.5-M7)} & \ding{52} & \textbf{86.4} & \textbf{74.2} & \textbf{42.3} & \textbf{29.2} & \textbf{31.6} & \textbf{52.4} \\
      \rowcolor{gray!30}
      \textsc{CIR~(Qwen3-NT8)} & \ding{52} & 93.2 & 91.6 & 61.5 & 46.3 & 38.8 & 66.3 \\
      \bottomrule
      \end{tabular}
      \label{tab:main_results}
\end{table}

\paratitle{Code-integrated RL outperforms conventional methods.}
As evidenced by Table~\ref{tab:main_results}, our proposed method (\textsc{CIR}) achieves consistent and superior performance across all evaluation benchmarks, attaining an average accuracy of 52.4\% that surpasses all baseline models. The improvements are particularly notable when compared to conventional RL approaches: \textsc{CIR} outperforms \textsc{Eurus-2-7B-PRIME} (37.6\%) by 14.8 percentage points and \textsc{Oat-Zero-7B} (39.3\%) by 13.1 percentage points. Even when compared to the recent tool-augmented RL method \textsc{ToRL-7B} (50.5\%), our approach maintains a clear performance advantage. This enhanced capability stems from our integration of an external code interpreter, which enables more effective code-augmented reasoning during the reinforcement learning process, ultimately leading to substantial gains in mathematical reasoning performance.


\paratitle{Code-integrated RL achieves superior performance on more challenging benchmarks.}
Code-integrated RL demonstrates significantly greater improvements on more challenging benchmarks. On AIME24, for example, \textsc{CIR} achieves 42.3\%, a substantial increase compared to \textsc{Eurus-2-7B-PRIME} (18.8\%) and \textsc{Oat-Zero-7B} (31.3\%). This trend persists on AIME25, where our method achieves 29.2\% accuracy—nearly doubling \textsc{Eurus-2-7B-PRIME}'s performance (15.4\%) and outperforming \textsc{Oat-Zero-7B} (10.4\%) by 18.8 points. Most notably, on the particularly challenging OlymMATH benchmark, \textsc{CIR} maintains strong performance of 31.6\%, demonstrating the method's exceptional capability in handling complex problems. These results underscore how code-integrated reasoning enhances the model's capacity to solve intricate mathematical problems through more sophisticated solution strategies.

Note that in Table~\ref{tab:main_results}, we also include results for the Qwen3-8B model trained in non-thinking mode, which achieves the highest performance across all benchmarks. However, since Qwen3-8B was specifically designed to support both thinking and non-thinking modes, direct comparison with other methods trained exclusively on non-thinking models may not be appropriate. We therefore present its results in gray for reference purposes only, not for direct comparison.


\subsubsection{Learning Dynamics}

\begin{figure}[htbp] 
\centering 
\subfigure[Training Reward]{\label{fig:reward}
\includegraphics[width=0.40\linewidth]{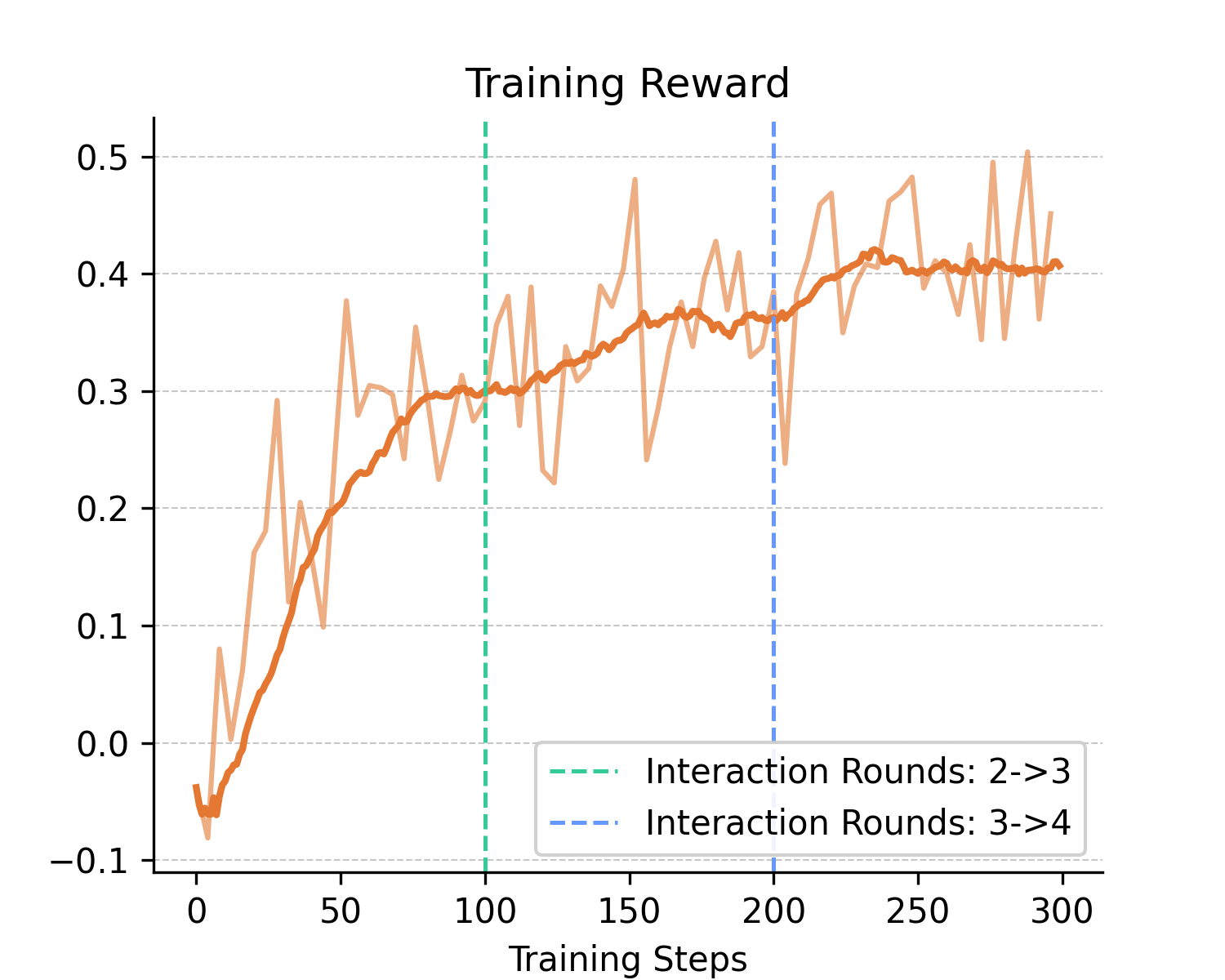}}
\hspace{-2.5em}
\subfigure[Response Length]{\label{fig:length}
\includegraphics[width=0.40\linewidth]{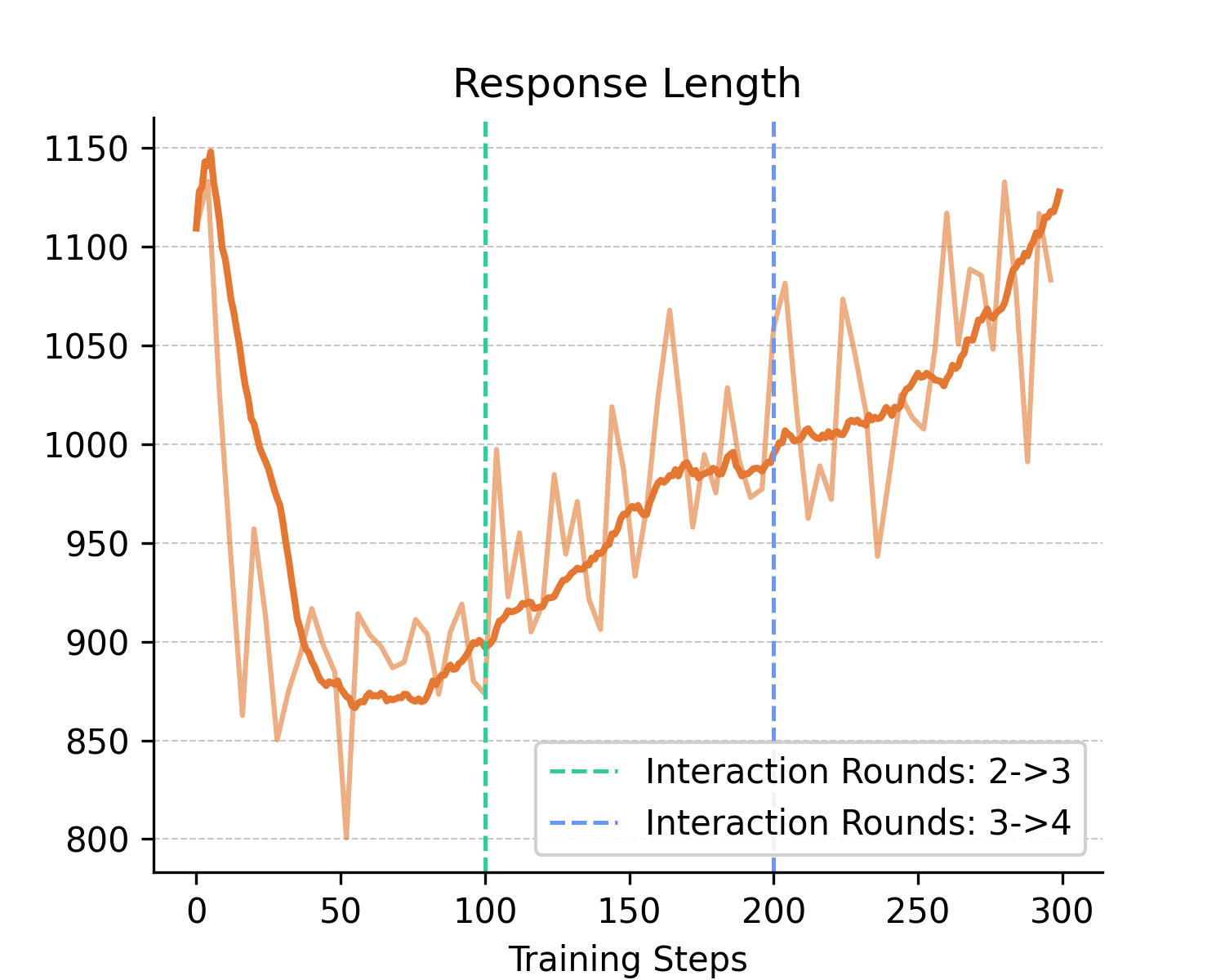}}
\caption{Training reward and average response length during the training process.}
\label{fig:reward_length}
\end{figure}

\begin{figure} 
\centering 
\subfigure[AIME2024]{\label{fig:aime24}
\includegraphics[width=0.40\linewidth]{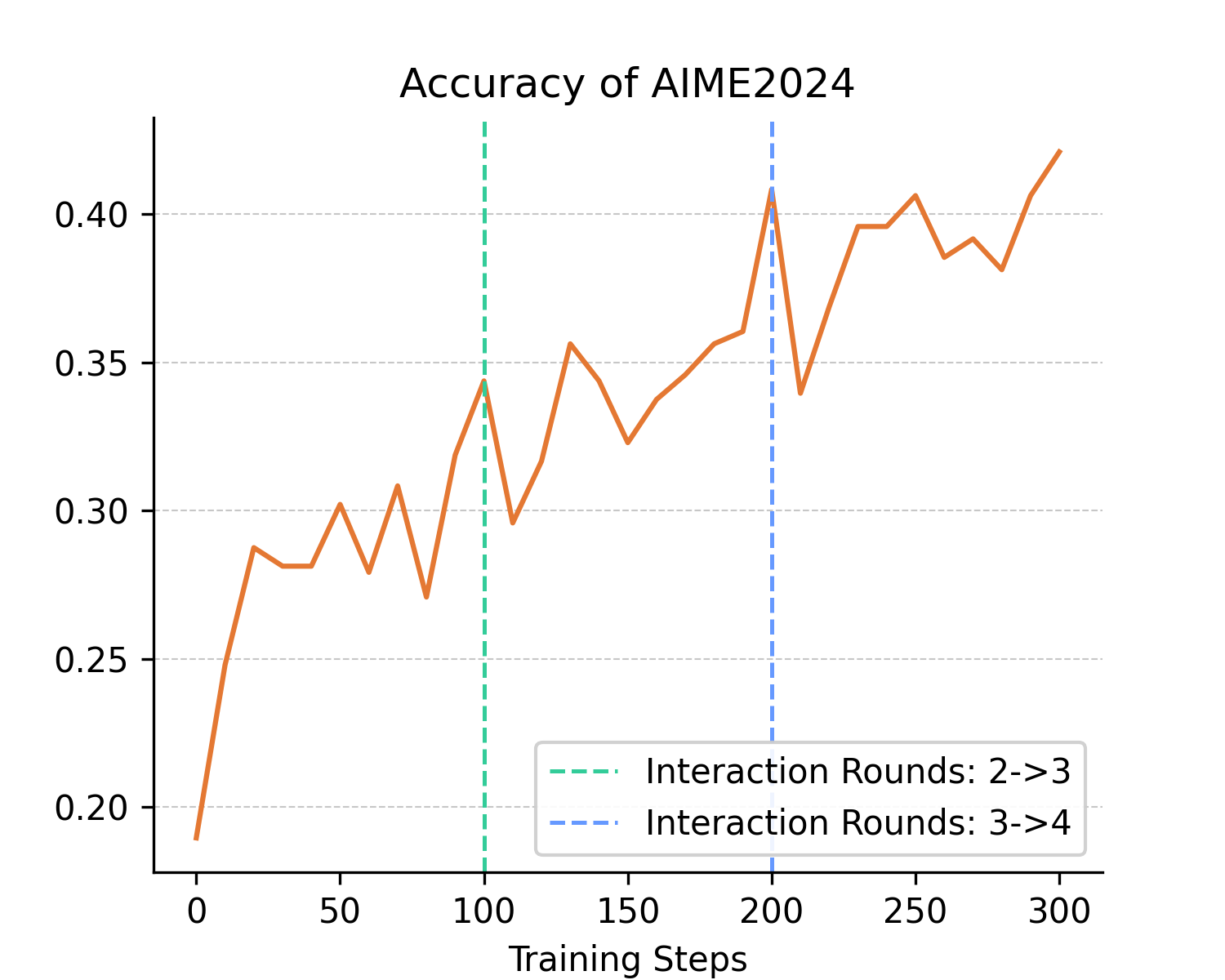}}
\hspace{-2.5em}
\subfigure[MATH500]{\label{fig:math500}
\includegraphics[width=0.40\linewidth]{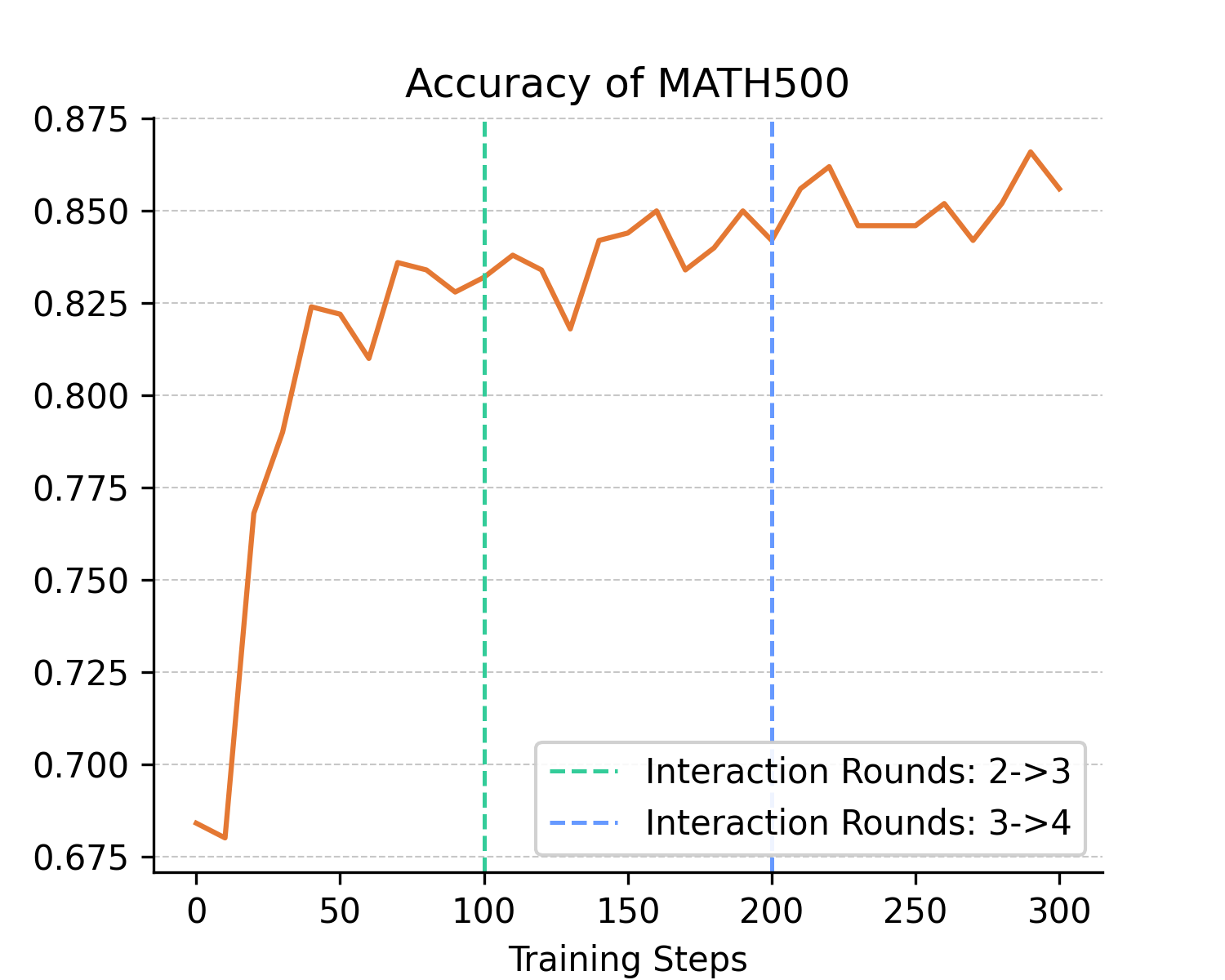}}
\caption{The test accuracy on AIME2024 and MATH500.}
\label{fig:testset}
\end{figure}

\begin{figure} 
\centering 
\subfigure[Average Code Number]{\label{fig:avg-code_num}
\includegraphics[width=0.36\linewidth]{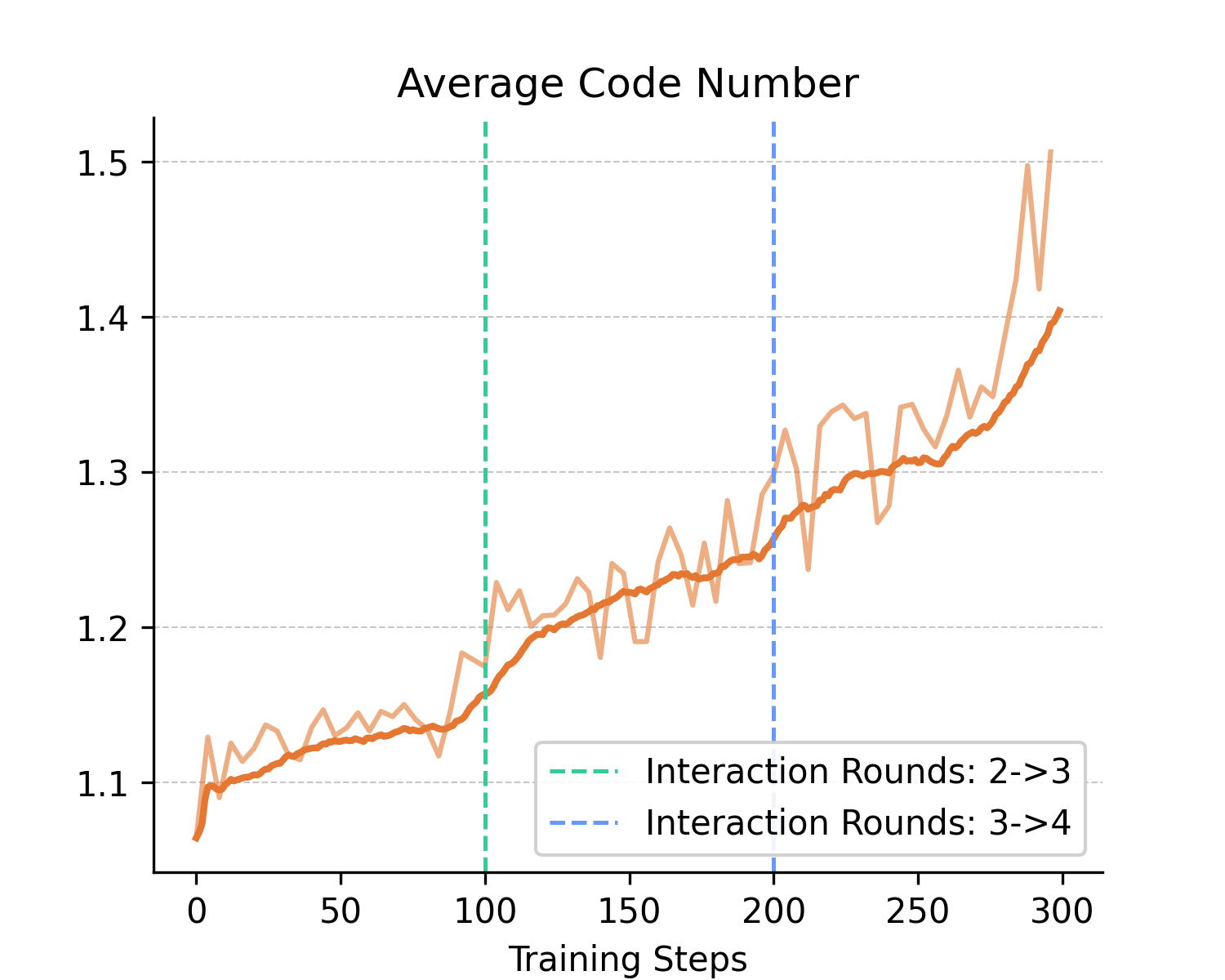}}
\hspace{-2.5em}
\subfigure[Average Code Ratio]{\label{fig:code_ratio}
\includegraphics[width=0.36\linewidth]{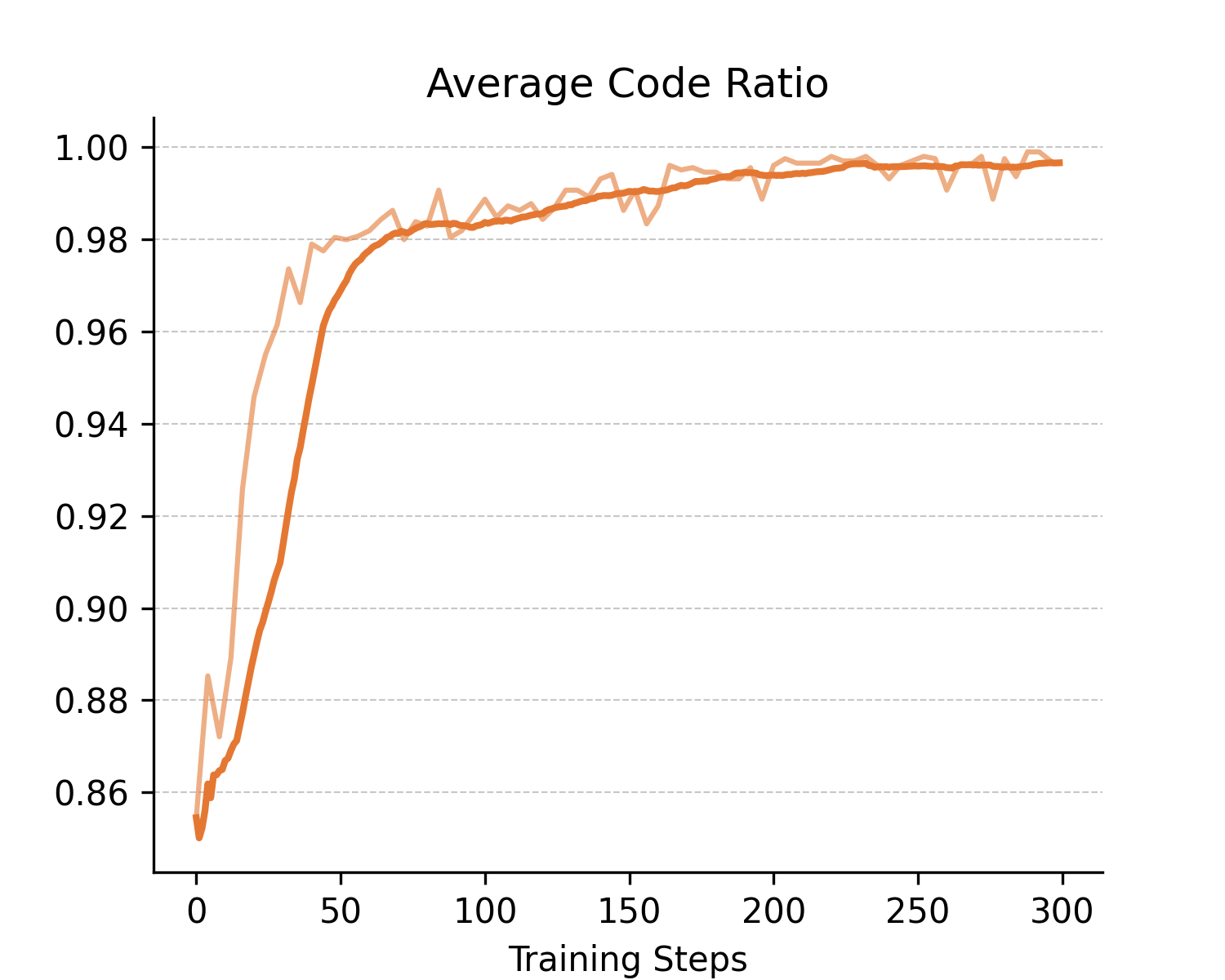}}
\hspace{-2.5em}
\subfigure[Code Pass Rate]{\label{fig:pass_rate}
\includegraphics[width=0.36\linewidth]{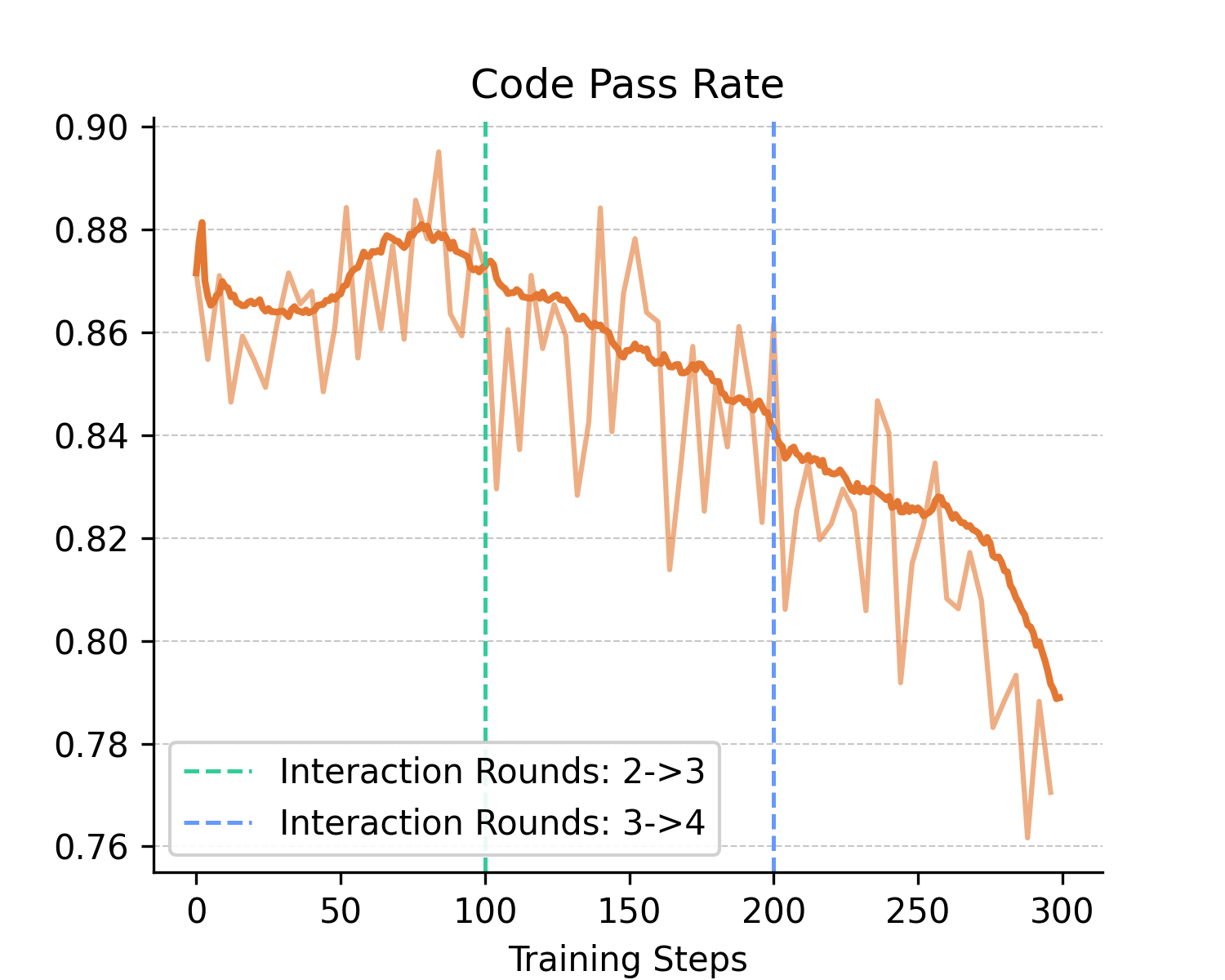}}
\caption{The average code generation number, code generation ratio and code pass rate during the training process.}
\label{fig:code_behavior}
\end{figure}

To analyze training stability more thoroughly, we present the learning dynamics of \textsc{CIR} in Figure~\ref{fig:reward_length}, Figure~\ref{fig:testset}, and Figure~\ref{fig:code_behavior}. During training, the interaction budget is progressively expanded from 2 to 3 and then from 3 to 4. Figure~\ref{fig:reward_length} reveals two key observations: (1) the model's reward increases steadily, demonstrating its ability to adapt to and benefit from the growing budget; and (2) the response length drops significantly in early stages, indicating a transition from traditional text-based reasoning to a more efficient tool-invocation approach. As training progresses, the model's demand for interactions grows, and the expanded budget accommodates this need, resulting in increased response lengths—a change primarily driven by the higher number of interactions. Figure~\ref{fig:testset} further corroborates this trend on the AIME2025 and MATH500 test sets, showing a clear positive correlation between budget expansion and improved model performance. 
We find that the Qwen2.5-Math backbone adapts well to code-integrated RL training. This is likely because mainstream math models are typically exposed to significant amounts of code data during training, enabling them to generate code with relative stability from the outset.


Furthermore, as shown in Figure~\ref{fig:code_behavior}, the model generates an average of more than one code snippet per response starting from the early training stages, with a code generation rate exceeding 85\%. As training progresses and the budget gradually expands, the model's reliance on code for problem-solving increases, reaching nearly 100\% code generation in later stages. 
However, the code pass rate exhibits a more complex pattern: it shows an initial upward trend in the first 100 steps, followed by a steady decline. This decline becomes more pronounced as the interaction budget increases—a counterintuitive result that suggests a lower pass rate does not necessarily indicate poorer performance. We will analyze this phenomenon further in Section~\ref{subsec:passrate}.

\section{Mechanistic Analysis of Code-Integrated Reasoning}
\label{sec-analysis}

While code ingetration during reasoning has proven effective for solving mathematical problems, the underlying mechanisms behind its efficacy and its contribution to performance improvement remain understudied. In this section, we examine the role of code-integrated reasoning, analyzing both its fundamental benefits and the specific mechanisms by which it enhances model performance in mathematical tasks.


\subsection{Impact on Model's Capability Boundaries}

In terms of the learning paradigm, reinforcement learning (RL) incentivizes models to explore and improve based on supervision signals from reward models. As such, recent studies~\cite{yue2025does,gandhi2025cognitive} suggest that RL does not extend a model's capacity boundaries, as it primarily relies on self-generated trajectories for training. In contrast, code-integrated reasoning typically invokes external code interpreters to execute generated code, a process fundamentally distinct from the reasoning patterns of standard text-based language models. We thus speculate that code-integrated reasoning may significantly expand the model's capability boundaries. To investigate this hypothesis, we conduct an empirical study.


\begin{figure}[htbp]
\begin{promptbox}[Prompt with Code-Integration]{cyan}
\texttt{Please solve the following problem step by step. During your reasoning process, if needed, you can choose to write python code to enhance your reasoning. The code executor will run your code and provide the execution results back to you to support your reasoning process. Please put the final answer within \textbackslash boxed\{\}.}
\end{promptbox}
\caption{The prompt triggering the model to utilize code-integrated reasoning.}
\label{fig:prompt}
\end{figure}

\begin{figure}[htbp] 
\centering 
\subfigure[AIME2024]{\label{fig:passk_24}
\includegraphics[width=0.4\linewidth]{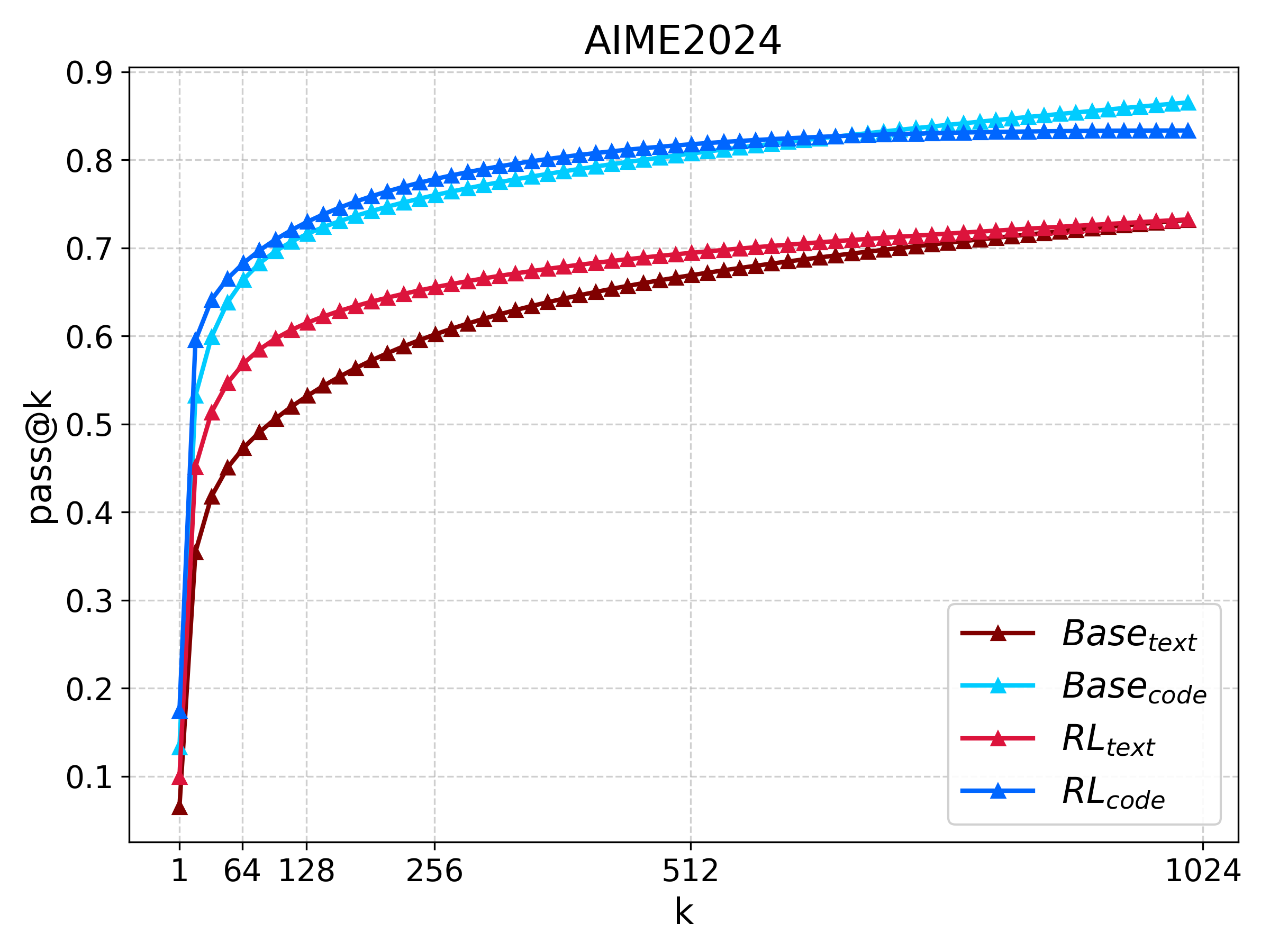}}
\hspace{0.01\linewidth}
\subfigure[AIME2025]{\label{fig:passk_25}
\includegraphics[width=0.4\linewidth]{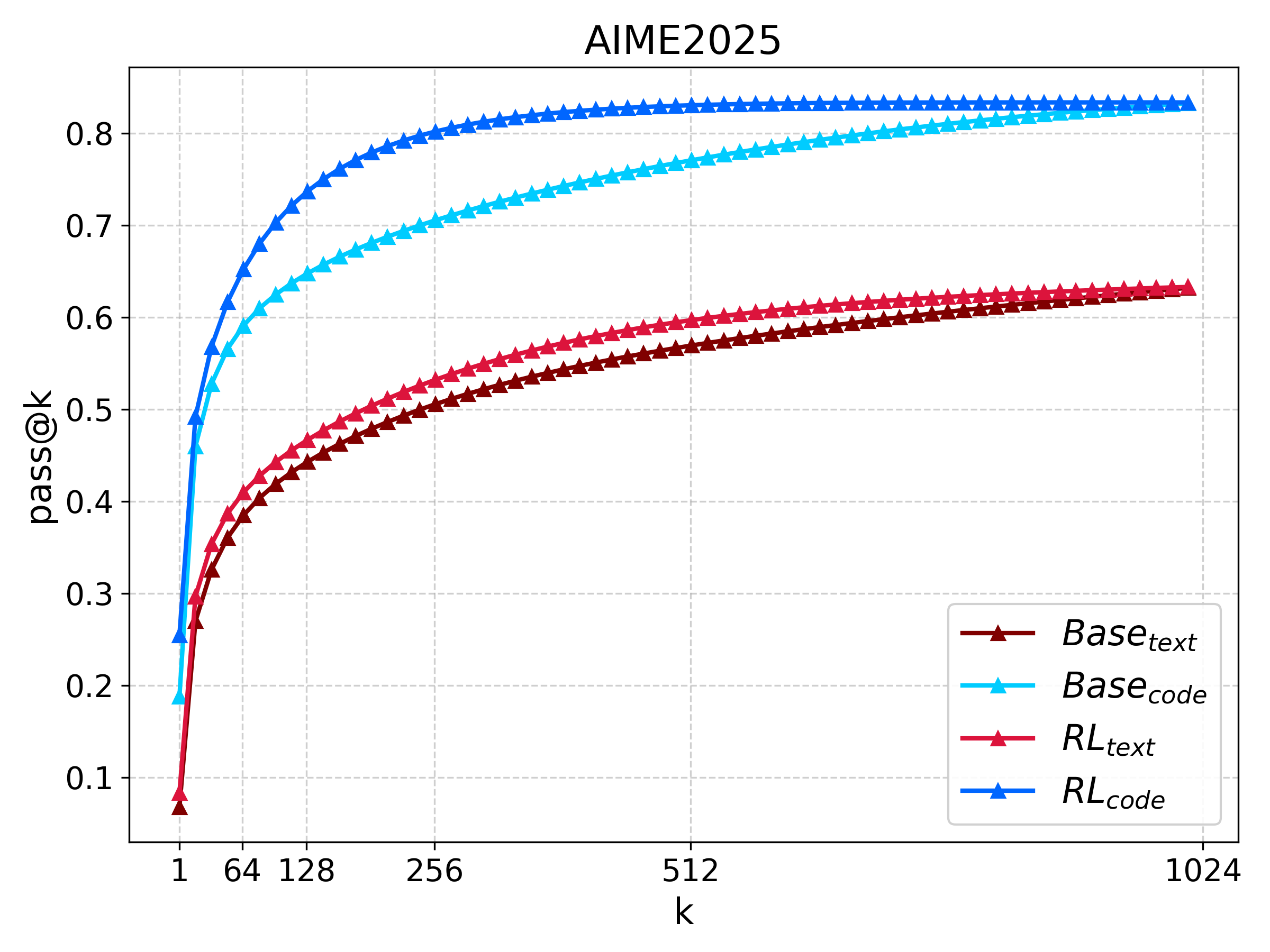}}
\vfill
\subfigure[OlymMATH-Easy]{\label{fig:passk_easy}
\includegraphics[width=0.4\linewidth]{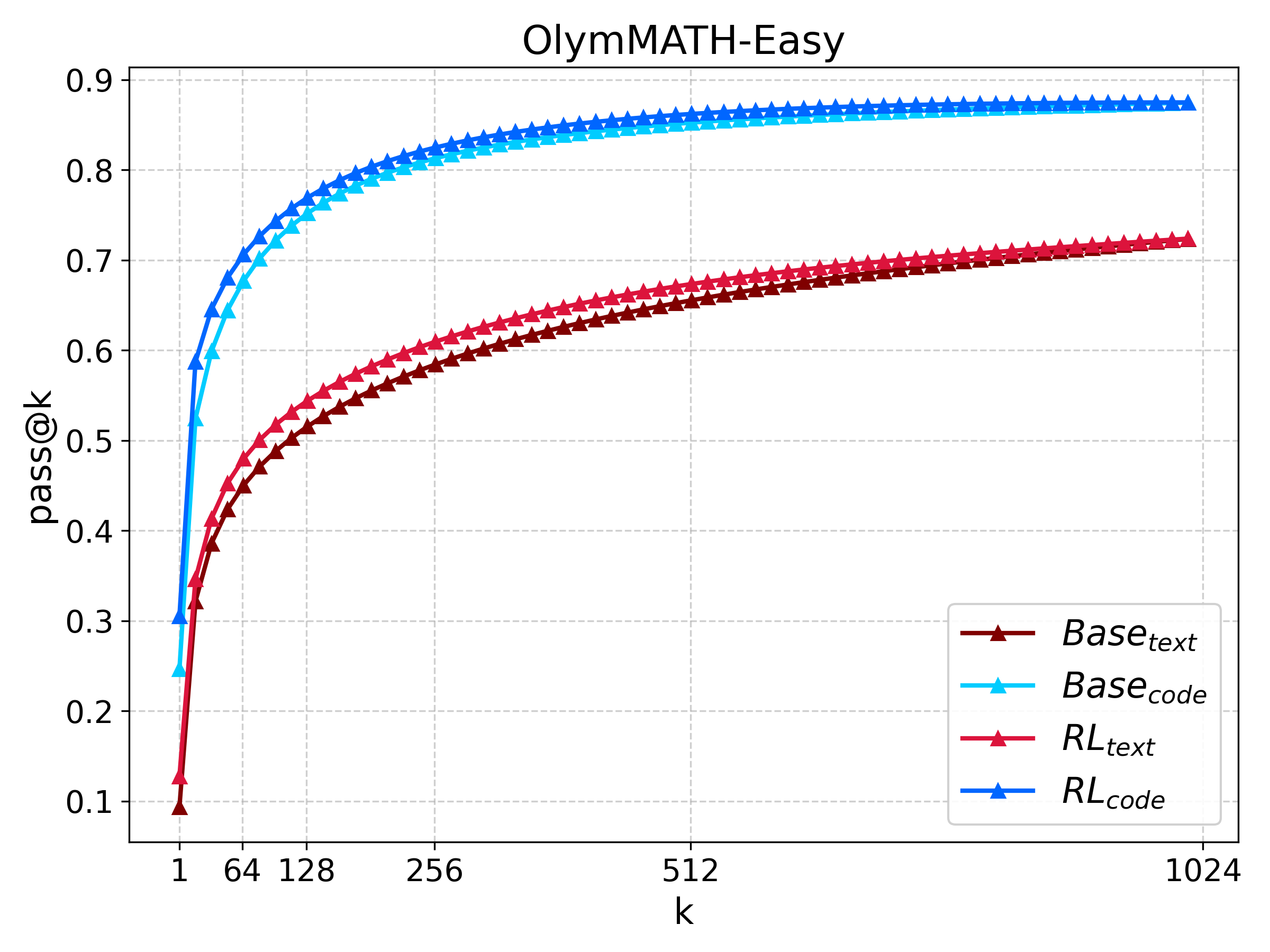}}
\hspace{0.01\linewidth}
\subfigure[OlymMATH-Hard]{\label{fig:passk_hard}
\includegraphics[width=0.4\linewidth]{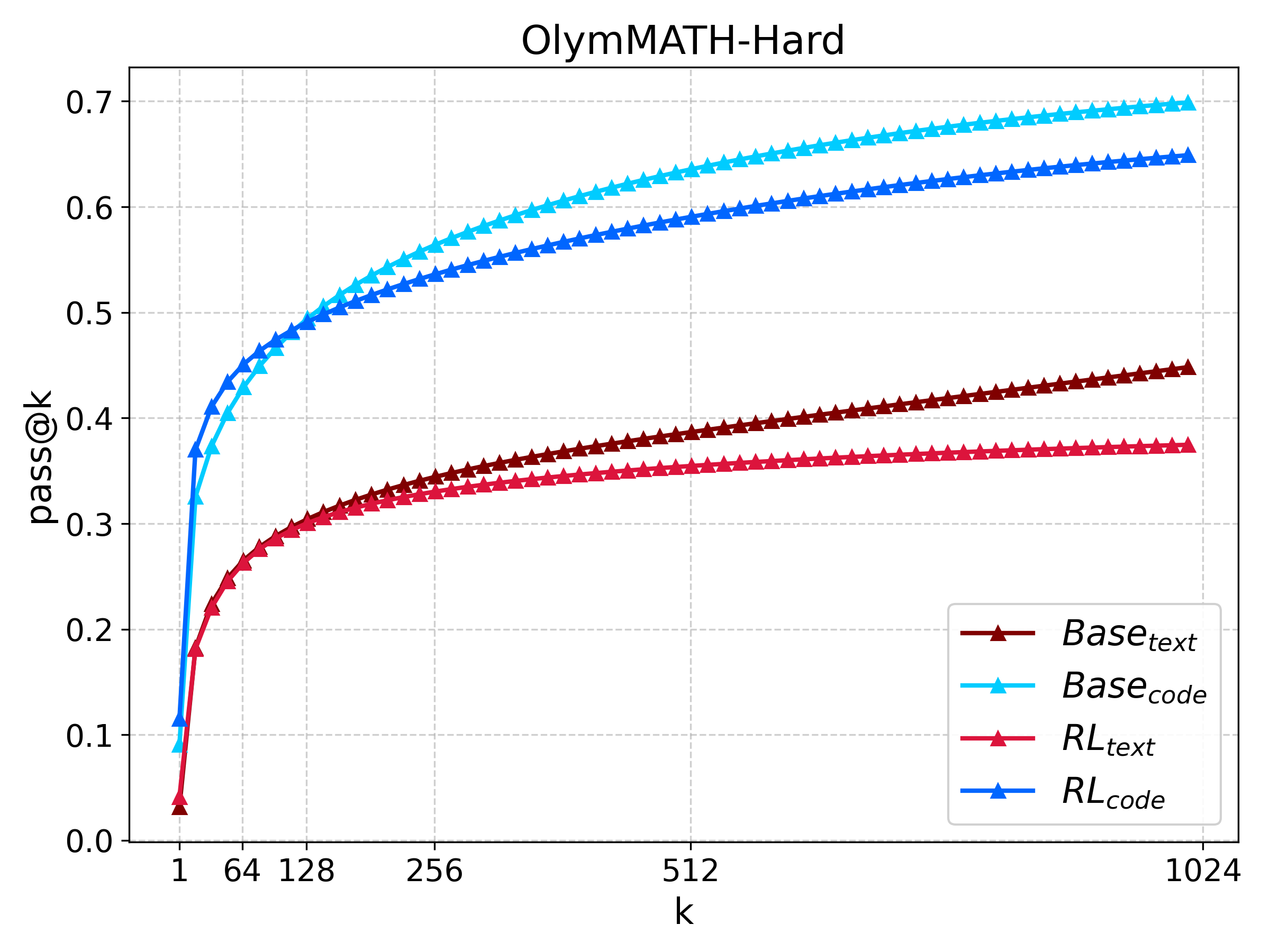}}
\caption{\textsc{Pass@$k$} Accuracy on AIME2024, AIME2025, OlymMATH-Easy and OlymMATH-Hard.}
\label{fig:passk}
\end{figure}

\paratitle{Evaluation settings.}
To investigate the impact of code-integrated reasoning on the model's capability boundaries, we test four baseline models, which are all derived from Qwen2.5-Math-7B, on the AIME2024, AIME2025, OlymMATH-Easy, and OlymMATH-Hard benchmarks, using the \textsc{Pass@$k$} metric. For all experiments, we use a temperature of 0.6 and a top-p value of 0.95, allowing for a maximum generation of 16,384 tokens. The four baselines are as follows:

$\bullet$ \emph{Base$_{text}$}: The base model, utilizing standard text-based reasoning.

$\bullet$ \emph{Base$_{code}$}: The base model, utilizing code-integrated reasoning via the special prompt, which is shown in Figure~\ref{fig:prompt}.

$\bullet$ \emph{RL$_{text}$}: The model trained with tool-augmented RL, utilizing text-based reasoning.

$\bullet$ \emph{RL$_{code}$}: The model trained with tool-augmented RL, utilizing code-integrated reasoning.

\paratitle{Code-integrated reasoning can expand model's capability boundaries.}
As illustrated in Figure~\ref{fig:passk}, the \textsc{Pass@$k$} performance of the model utilizing code-integrated reasoning is significantly higher than that of text-based reasoning across all four benchmarks. Specifically, the base model with specially designed tool-use prompts (\ie \emph{Base$_{code}$}) achieves the highest \textsc{Pass@$k$} performance when $k$ reaches a large value. This finding aligns with observations from recent studies~\cite{yue2025does}, where the base model demonstrates superior performance in correct answer coverage than the RL model as the rollout number increases. 
Another interesting observation is that for code-integrated reasoning models, removing the execution of the code interpreter results in a significant performance drop (\ie RL$_{text}$). This suggests that, with RL training, code generation and execution become integrated as fundamental model functionalities. If the execution functionality is discarded, the model struggles to adapt, as it inherently relies on code-integrated reasoning.


\subsection{Comparison between Code-Integrated Reasoning and Long-CoT Reasoning}

\begin{table}[htbp]
    \centering
    \small
    \setlength\tabcolsep{2.4pt}
    \caption{The accuracy and average response length on AIME2024, AIME2025 and OlymMATH between long-CoT and code-integrated reasoning models.}
      \begin{tabular}{lccccccccc}
      \toprule
      \multirow{2.5}*{\textbf{Models}} & \multirow{2.5}*{\textbf{Mode}} & \multicolumn{2}{c}{\textbf{AIME2024}} & \multicolumn{2}{c}{\textbf{AIME2025}} & \multicolumn{2}{c}{\textbf{OlymMATH}} & \multicolumn{2}{c}{\textbf{Avg}}\\
      \cmidrule(r){3-4}\cmidrule(r){5-6}\cmidrule(r){7-8}\cmidrule(r){9-10}
      & & Acc & \#Tokens & Acc & \#Tokens & Acc & \#Tokens & Acc & \#Tokens\\
      \midrule
      \textsc{Qwen3-8B (Thinking-10K)} & LCoT & 44.6 & 9296 & 32.5 & 9604 & 21.9 & 10216 & 33.0 & \underline{9705}\\
      \textsc{Distill-Qwen-7B} & LCoT & 51.3 & 12974 & 40.6 & 13479 & 42.2 & 15041 & \textbf{44.7} & 13831\\
      \textsc{CIR~(Qwen2.5-M7)} & CIR & 42.3 & 1603 & 29.2 & 1843 & 31.6 & 1733 & \underline{34.4} & \textbf{1726}\\
      \bottomrule
      \end{tabular}
      \label{tab:response_length}
\end{table}

\begin{table*}[ht]
    \small
    \centering
    \begin{tabular}{ll}
        \toprule
        \multicolumn{1}{l}{\begin{tabularx}{0.13\textwidth}{@{}X@{}}
            \textbf{Problem}
        \end{tabularx}} & \multicolumn{1}{l}{\begin{tabularx}{0.8\textwidth}{@{}X@{}}
            Find the largest possible real part of $(75+117i)z+\frac{96+144i}{z}$, where $z$ is a complex number with $|z|=4$.
        \end{tabularx}} \\
        \midrule
        \multicolumn{1}{l}{\begin{tabularx}{0.13\textwidth}{@{}X@{}}
            \textbf{\textsc{CIR~(Qwen2.5-M7)}}
        \end{tabularx}} & \multicolumn{1}{l}{\begin{tabularx}{0.8\textwidth}{@{}X@{}}
            Let's solve the problem step-by-step using reasoning and Python code.\\\textcolor{red}{\#\#\# Step-by-Step Reasoning}:
            
            1. **Given Expression and Constraints:** ...
            
            2. **Substitute $z = 4e^{i\theta}$:** ...
            
            3. **Expand and Separate Real and Imaginary Parts:** ...
            
            So the real part of the expression is $ 324\cos\theta - 432\sin\theta$. We need to maximize this expression. To find the maximum value of the real part, we need to recognize it as a function of the form $a\cos\theta + b\sin\theta$, and its maximum value is given by $\sqrt{a^2 + b^2}$.
            
            \textcolor{red}{\#\#\# Python Code Implementation}:
            \begin{tcolorbox}[colback=gray!20, colframe=codegray!50, boxrule=0pt,fontupper=\ttfamily]
            
            ```python
            
            import sympy as sp
            
            theta = sp.symbols('theta', real=True)
            
            real\_part = 324 * sp.cos(theta) - 432 * sp.sin(theta)
            
            max\_real\_part = sp.sqrt(324**2 + 432**2)
            
            print(max\_real\_part)
            
            ```
            
            ```output
            
            540
            
            ```
            \end{tcolorbox} 
            The maximum value of the real part of the expression $(75 + 117i)z + \frac{96 + 144i}{z}$ when $|z| = 4$ is $\boxed{540}$.
        \end{tabularx}} \\
        \midrule
        \multicolumn{1}{l}{\begin{tabularx}{0.13\textwidth}{@{}X@{}}
            \textbf{\textsc{DeepSeek-R1-Distill-Qwen-7B}}
        \end{tabularx}} & \multicolumn{1}{l}{\begin{tabularx}{0.8\textwidth}{@{}X@{}}
       <think>\\Okay, so I have this problem: I need to find the largest possible real part of the expression $(75 + 117i)z + \frac{96 + 144i}{z}$, where $z$ is a complex number with $|z| = 4$. Hmm, okay. Let me try to figure this out step by step.\\...\\ \textcolor{red}{Wait, hold on, actually no}. The real part is the sum of the real parts of each term, and similarly for the imaginary part.\\...\\ \textcolor{red}{Wait, let me double-check because sometimes there could be a mistake}. Wait, $81^2$ is indeed 6561, $108^2$ is 11664, adding together gives 18225, whose square root is 135, which seems correct.\\...\\ \textcolor{red}{Wait, so perhaps I could have approached this problem using complex conjugates or other complex analysis techniques}. Let me try that as an alternative approach to confirm.\\...\\Therefore, whether I take the parametrization approach or the Cauchy-Schwarz, I get 540.\\</think>\\Therefore, the largest possible real part is $\boxed{540}$.
        \end{tabularx}} \\
        \bottomrule
    \end{tabular}
    \caption{The comparison of the responses from long-CoT model and code-integrated reasoning model.}
    \label{case_study_qa}
\end{table*}

Above, we have carefully examined the effectiveness of code-integrated reasoning. We now compare it with standard long-chain-of-thought (long-CoT) reasoning that does not use external tools. For this comparison, we select two reference models: \textsc{Qwen3-8B (Thinking-10K)} and \textsc{DeepSeek-R1-Distill-Qwen-7B}. To ensure evaluation efficiency, we limit the output window of \textsc{Qwen3-8B} to 10K tokens. Table~\ref{tab:response_length} presents the comparison results, including accuracy performance and response length.

As shown in Table~\ref{tab:response_length}, our model achieves comparable accuracy while significantly reducing response length. Specifically, it attains nearly 80\% of the accuracy of \textsc{DeepSeek-R1-Distill-Qwen-7B} using less than 15\% of its response length, and matches the performance of \textsc{Qwen3-8B (Thinking-10K)} with under 20\% of the response length. This demonstrates that code-integrated reasoning enables models to achieve competitive performance with much greater conciseness.


To investigate why code-integrated reasoning produces more concise responses, we manually inspect the generated outputs and present a representative example in Table~\ref{case_study_qa}. We find that code-integrated reasoning typically begins with a brief yet complete overview of the solution approach before generating executable code, which explains its conciseness. In contrast, Long-CoT models tend to continuously decompose problems, switch strategies, and engage in self-reflection throughout the reasoning process, resulting in significantly longer responses.



The differences between these reasoning modes can be summarized as follows:

$\bullet$ \textbf{Code-integrated reasoning  enables efficient, precise computation through programming.}
By integrating executable code in responses, this approach substantially improves algorithmic precision and minimizes ambiguity. It leverages the structured efficiency of programming languages to deliver fast, deterministic outputs while bypassing verbose reasoning steps.

$\bullet$ \textbf{Long-CoT reasoning mimics exploratory, human-like thought processes.}
Rather than following rigid procedures, long-CoT reasoning simulates iterative reflection—revising hypotheses, exploring alternatives, and refining reasoning paths. This often results in longer responses as the model evaluates multiple approaches before converging to a solution.

Note that this conclusion is based on the fact that our model is trained from a base model rather than a reasoning model. 
It would also be meaningful to conduct similar tool-augmented RL training on long-CoT reasoning models. However, since these models already produce responses of considerable length, the associated training cost would be significantly higher than in our current experiments. We leave this investigation for future work.

\subsection{Effect of Non-Executable Code in Code-Integrated Reasoning}
\label{subsec:passrate}
Translating problem-solving steps into codes and executing them via an external interpreter can significantly enhance the model’s accuracy. While producing correct and executable code is clearly beneficial, the impact of generating non-executable code on problem-solving performance remains understudied. 

To investigate this, we analyze the code pass rate for both correct and incorrect responses generated by our model trained with tool-augmented RL on 60 problems from AIME2024 and AIME2025, with 16 samples generated per problem. As illustrated in Table~\ref{tab:error}, about one-third of the correct responses still contain code that fails to execute, while more than one-third of the incorrect responses include fully executable code. Based on these results and further case analysis, we draw the following conclusions.

$\bullet$ \textbf{Executable but logically misaligned code might limit further improvements in model performance.}
Executable code often provides the model with feedback that appears valid, but these formally correct results can prematurely terminate the reasoning process and lead to incorrect answers. As shown in Table~\ref{tab:error}, 36.0\% of incorrect responses stem specifically from this type of error, where executable but logically flawed code misleads the model. This demonstrates that code executability does not reliably indicate response quality. In fact, correctly executed code may be more likely to mislead the model, as it tends to accept such feedback without further verification.


$\bullet$ \textbf{Error feedback from non-executable code may also benefit code-integrated reasoning.}
Non-executable code generates informative error feedback that the model cannot produce independently, compelling it to reflect on and revise its code. This process increases the likelihood of developing both executable and logically sound solutions, ultimately improving accuracy. As shown in Table~\ref{tab:error}, 39.4\% of correct responses include non-executable code, suggesting that such code may still contribute to valid reasoning paths. Furthermore, the counter-intuitive declining pass rate in Figure~\ref{fig:code_behavior} suggests that during training, the increased frequency of interactions with the code interpreter may result from more extensive code revisions.
This observation is further supported by the contrast between the \emph{Full Pass Rate} (60.6\%) and \emph{Final Pass Rate} (65.5\%) for correct responses in Table~\ref{tab:error}, where approximately 5\% of responses represent successfully revised code that lead to correct answers.



\begin{table}[htbp]
    \centering
    \small
    \setlength\tabcolsep{2.4pt}
    \caption{The average {code generation number}, {full pass rate}, {error rate} and {final pass rate} for correct and incorrect responses, where \emph{Avg Code Num} refers to the average number of code snippets in a response, \emph{Full Pass Rate} denotes the proportion of responses in which all code executions succeed, \emph{Error Rate} means the proportion of responses that contain at least one segment of non-executable code, and \emph{Final Pass Rate} measures the percentage of responses whose final generated code runs without execution errors. The number within the bracket indicates the number of responses.}
      \begin{tabular}{lcccc}
      \toprule
       &  \textbf{Avg Code Num} & \textbf{Full Pass Rate} & \textbf{Error Rate} & \textbf{Final Pass Rate}\\
      \midrule
      \textsc{correct response (345)}  & 1.47 &  60.6\%  & 39.4\% & 65.5\%\\
      \textsc{incorrect response (615)}  & 2.11 &  36.0\% & 64.0\% & 40.2\%\\

      \bottomrule
      \end{tabular}
      \label{tab:error}
\end{table}

\begin{table*}[ht]
    \small
    \centering
    \begin{tabular}{ll}
        \toprule
        \multicolumn{1}{l}{\begin{tabularx}{0.13\textwidth}{@{}X@{}}
            \textbf{Problem}
        \end{tabularx}} & \multicolumn{1}{l}{\begin{tabularx}{0.8\textwidth}{@{}X@{}}
            Let $ABCD$ be a tetrahedron such that $AB=CD= \sqrt{41}$, $AC=BD= \sqrt{80}$, and $BC=AD= \sqrt{89}$. There exists a point $I$ inside the tetrahedron such that the distances from $I$ to each of the faces of the tetrahedron are all equal. This distance can be written in the form $\frac{m \sqrt n}{p}$, where $m$, $n$, and $p$ are positive integers, $m$ and $p$ are relatively prime, and $n$ is not divisible by the square of any prime. Find $m+n+p$.
        \end{tabularx}} \\
        \midrule
        \multicolumn{1}{l}{\begin{tabularx}{0.13\textwidth}{@{}X@{}}
            \textbf{Solution}
        \end{tabularx}} & \multicolumn{1}{l}{\begin{tabularx}{0.8\textwidth}{@{}X@{}}
            Let's begin by analyzing the given tetrahedron \(ABCD\) with the given edge lengths:\\
            
            To find the point \(I\) inside the tetrahedron such that the distances from \(I\) to each of the faces of the tetrahedron are all equal, we need to find the incenter of the tetrahedron.\\
            The incenter of a tetrahedron is the point equidistant from all four faces, and this distance is the inradius of the tetrahedron.\\
            
            To find the inradius, we need to use the formula for the inradius \(r\) of a tetrahedron:
            \[
            r = \frac{3V}{A}
            \]
            where \(V\) is the volume of the tetrahedron and \(A\) is the total surface area of the tetrahedron.\\
            
            First, we will calculate the surface area of the tetrahedron. Each face of the tetrahedron is a triangle. Let's calculate the area of each triangular face using Heron's formula.\\
            
            For a triangle with sides \(a\), \(b\), and \(c\), the semi-perimeter \(s\) is given by:
            \[
            s = \frac{a+b+c}{2}
            \]
            and the area \(K\) is given by:
            \[
            K = \sqrt{s(s-a)(s-b)(s-c)}
            \]
            
            We will calculate the area of each face:\\
            ... \\
            
            
            
            \textcolor{red}{The total surface area \(A\) of the tetrahedron is the sum of the areas of the four faces:
            \[
            A = K_{ABC} + K_{ABD} + K_{ACD} + K_{BCD} = K_{ABC} + 3K_{ABD}
            \]}
            
            Next, we need to find the volume \(V\) of the tetrahedron. We can use the Cayley-Menger determinant for the volume of a tetrahedron:
            \[
            288 V^2 = 
            \begin{vmatrix} 
            0 & 1 & 1 & 1 & 1 \\ 
            1 & 0 & d_{12}^2 & d_{13}^2 & d_{14}^2 \\ 
            1 & d_{12}^2 & 0 & d_{23}^2 & d_{24}^2 \\ 
            1 & d_{13}^2 & d_{23}^2 & 0 & d_{34}^2 \\ 
            1 & d_{14}^2 & d_{24}^2 & d_{34}^2 & 0 
            \end{vmatrix}
            \]
            Using Python to compute the determinant:\\
            \begin{tcolorbox}[colback=gray!20, colframe=codegray!50, boxrule=0pt,fontupper=\ttfamily]
            ```python\\
            import sympy as sp
            \\
            \# Define squared distances\\
            ...
    
            \# Cayley-Menger matrix\\
            ...
            
            \# Determinant and volume
            
            ...\\
            ```\\
            ```output\\
            2.06039101378971\\
            ```
            \end{tcolorbox}
            The result is approximately:

            $r \approx 2.06039101378971$
            
            We attempt to match this to a simplified radical form:
            
            $r = \frac{8\sqrt{5}}{7}$
            
            Then:
            
            $m = 8, \quad n = 5, \quad p = 7 \Rightarrow m+n+p = \boxed{20}$
            
        \end{tabularx}} \\
        \bottomrule
    \end{tabular}
    \caption{A case of geometry problem. $K_{ABD} + K_{ACD} + K_{BCD} \neq 3K_{ABD}$. A minor mistake in complex framework leads to logical error in generated code, causing the final wrong answer.}
    \label{tab:geometry}
\end{table*}

\subsection{Performance Gain Analysis of Code-Integrated Reasoning}
While code-integrated reasoning can enhance mathematical reasoning, its benefits may vary across different problem types. To better understand these variations, we examine how the advantages of code-integrated reasoning differ among various categories of mathematical problems.
We evaluate three models, \emph{Base$_{text}$}, \emph{Base$_{tool}$} and \emph{RL$_{tool}$}, on four main categories of mathematical problems (algebra, geometry, combinatorics, and number theory) from AIME 2024 and OlymMATH.

\begin{figure} 
\centering 
\subfigure[AIME2024]{\label{fig:category_24}
\includegraphics[width=0.45\linewidth]{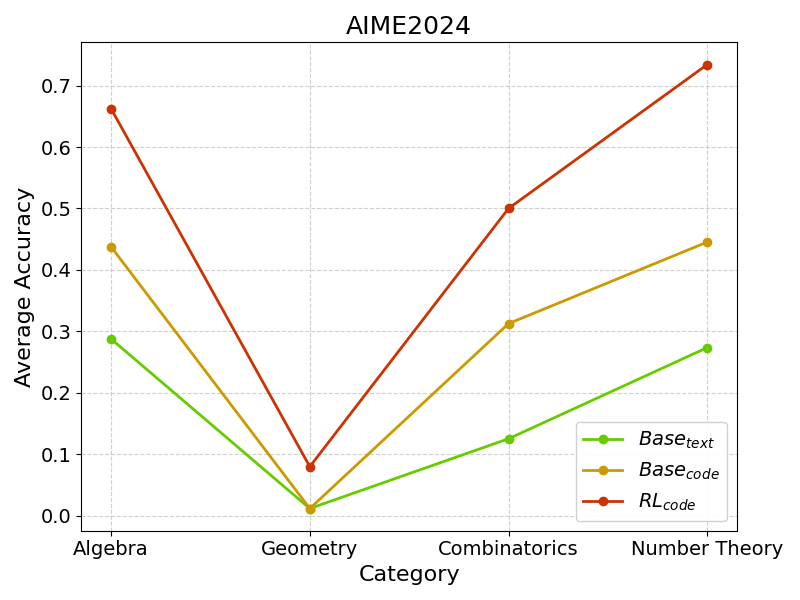}}
\hspace{0.01\linewidth}
\subfigure[OlymMATH]{\label{fig:category_OlymMATH}
\includegraphics[width=0.45\linewidth]{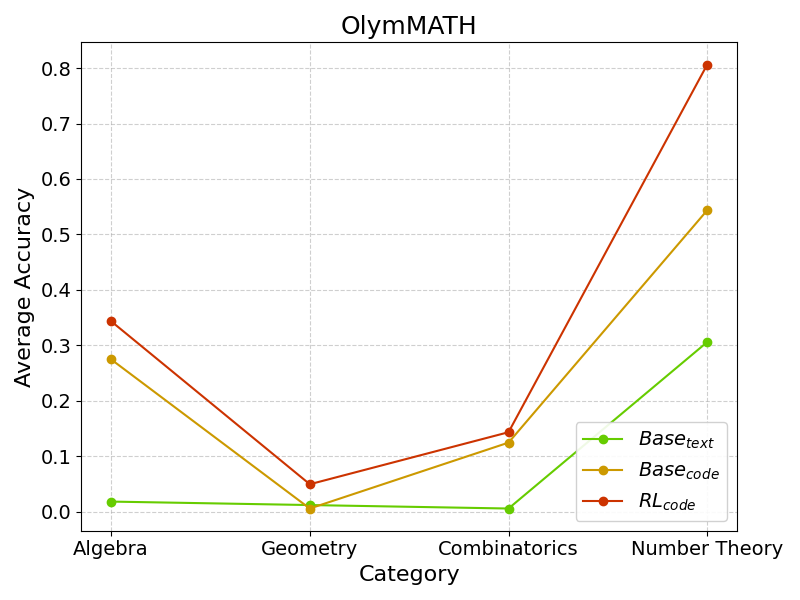}}
\caption{Accuracy of different categories on AIME2024 and OlymMATH.}
\label{fig:category}
\end{figure}

Figure~\ref{fig:category} demonstrates that the accuracy improvements from code integration vary significantly across problem types. Compared to pure text-based reasoning, code-integrated reasoning provides substantial gains for algebra, number theory, and combinatorics problems by offloading complex computations to precise external interpreters, thereby preventing error accumulation in lengthy text-based derivations. RL further amplifies this advantage. 
However, code integration offers only marginal benefits for geometry problems. Across both the AIME2024 and OlymMATH datasets, it yields minimal improvements in solution accuracy for this category. Even with RL, the performance gain remains significantly smaller than for other problem types. 

Table~\ref{tab:geometry} further illustrates a typical failure case for geometry problems. These problems demand abstract spatial reasoning and conceptual understanding from the outset—success hinges more on constructing correct geometric frameworks than computational precision. Consequently, code execution proves ineffective when the underlying logical reasoning is flawed.

\section{Conclusion}

In this paper, we present a systematic study on improving the training effectiveness of code-integrated reasoning models. We introduce a set of enhanced training strategies that jointly promote exploration and stabilize learning. With these improvements, our model (based on \textsc{Qwen2.5-MATH-7B}) achieves 42.3\% accuracy on AIME2024.
Beyond technical advances, we provide a detailed analysis of the role of code integration, showing that it extends the capacity boundaries of reasoning models and improves reasoning efficiency. Although our current implementation focuses on code execution, the framework is inherently extensible to other external tools. Future work will explore this generalization to further enhance model capabilities.

\bibliographystyle{unsrt}
\bibliography{ref.bib}

\end{document}